\documentclass[letterpaper, 10 pt, conference]{ieeeconf}  

\usepackage{graphicx}
\usepackage[misc]{ifsym}
\usepackage{url}

\usepackage{cite}

\usepackage{amsmath,amssymb,amsfonts}

\usepackage{algorithmic}

\usepackage{textcomp}

\usepackage{color}

\usepackage{multirow}

\usepackage{array}

\usepackage{soul}

\usepackage{bm}

\usepackage{ulem}
\usepackage{makecell}

\usepackage{subfig}

\usepackage{caption}
\usepackage{bbding}
\usepackage{subfig}

\IEEEoverridecommandlockouts                              


\title{\LARGE \bf
Pedestrian Motion Tracking by Using Inertial Sensors on the Smartphone
}

\author{Yingying Wang$^{*}$, Hu Cheng$^{*}$ and Max Q.-H. Meng\textsuperscript{\Envelope}, \textit{Fellow, IEEE} %
\thanks{$^{*}$Yingying Wang and Hu Cheng contributed equally to this work.}%
\thanks{\textsuperscript{\Envelope}Corresponding author, {\tt\small max.meng@ieee.org}}%
\thanks{Yingying Wang and Hu Cheng are with the Robotics, Perception and Artificial Intelligence Lab in the Department of Electronic Engineering, The Chinese University of Hong Kong, N.T., Hong Kong SAR, China.}%
\thanks{Max Q.-H. Meng is with the Department of Electronic and Electrical Engineering of the Southern University of Science and Technology in Shenzhen, China, on leave from the Department of Electronic Engineering, The Chinese University of Hong Kong, Hong Kong, and also with the Shenzhen Research Institute of the Chinese University of Hong Kong in Shenzhen, China.}%
}

\begin{document}

\maketitle
\thispagestyle{empty}
\pagestyle{empty}

\begin{abstract}

Inertial Measurement Unit (IMU) has long been a dream for stable and reliable motion estimation, especially in indoor environments where GPS strength limits. In this paper, we propose a novel method for position and orientation estimation of a moving object only from a sequence of IMU signals collected from the phone. Our main observation is that human motion is monotonous and periodic. We adopt the Extended Kalman Filter and use the learning-based method to dynamically update the measurement noise of the filter. Our pedestrian motion tracking system intends to accurately estimate planar position, velocity, heading direction without restricting the phone's daily use. The method is not only tested on the self-collected signals, but also provides accurate position and velocity estimations on the public RIDI dataset, i.e., the absolute transmit error is 1.28\textit{m} for a 59-second sequence.

\end{abstract}

\section{INTRODUCTION}

The need for accurate and reliable indoor localization in where the global positioning system (GPS) is not serviced sufficiently has continued to grow, such as customer navigation in supermarkets, augmented reality in public places. The smartphone can be a critical part of human indoor localization solution because of the wide range of sensors and its powerful computational ability \cite{langlois2017indoor}. Meanwhile, the cost of using the phone is negligible, because almost every one equips with one phone.

There are kinds of approaches used in indoor localization based on a phone. Traditionally, the signal between a transmitter and a receiver is used. For example, time of arrival (TOA) is used by using the signal's travel time from transmitter to receiver to calculate the distance  \cite{cheung2005multidimensional}. Angle of arrival (AOA) is the method that determines the direction of the incoming signal by exploiting and detecting phase difference among antennas \cite{shen2010fundamental}, \cite{zimmermann2012gsm}. There is also the distance information reflected by the strength of the received signal \cite{yang2015wifi}. These transmitted signal based methods require at least two types of equipment and the accuracy suffers due to the huge changeability of the environments, such as moving crowds. Fingerprinting method is conducted by the unique signal distribution at a specific place \cite{wang2016indoor}. Liang et al. \cite{liang2018indoor} combined multi-opportunistic signals, including Wifi, magnet, and lights, for offline mapping and online localization afterward. Camera-based localization has produced high precision accuracy and visual-inertial odometry combines visual features and IMU signals to produce a more robust result \cite{hesch2014camera}, \cite{qin2018vins}, such as Google Tango project \cite{WinNT}. These kinds of methods need the camera to be exposed, which is power consumed and privacy affected, and the accuracy is closely connected with the lights of the environment.

\begin{figure}[tbp]
	\centering 
	\includegraphics[width=2.2in]{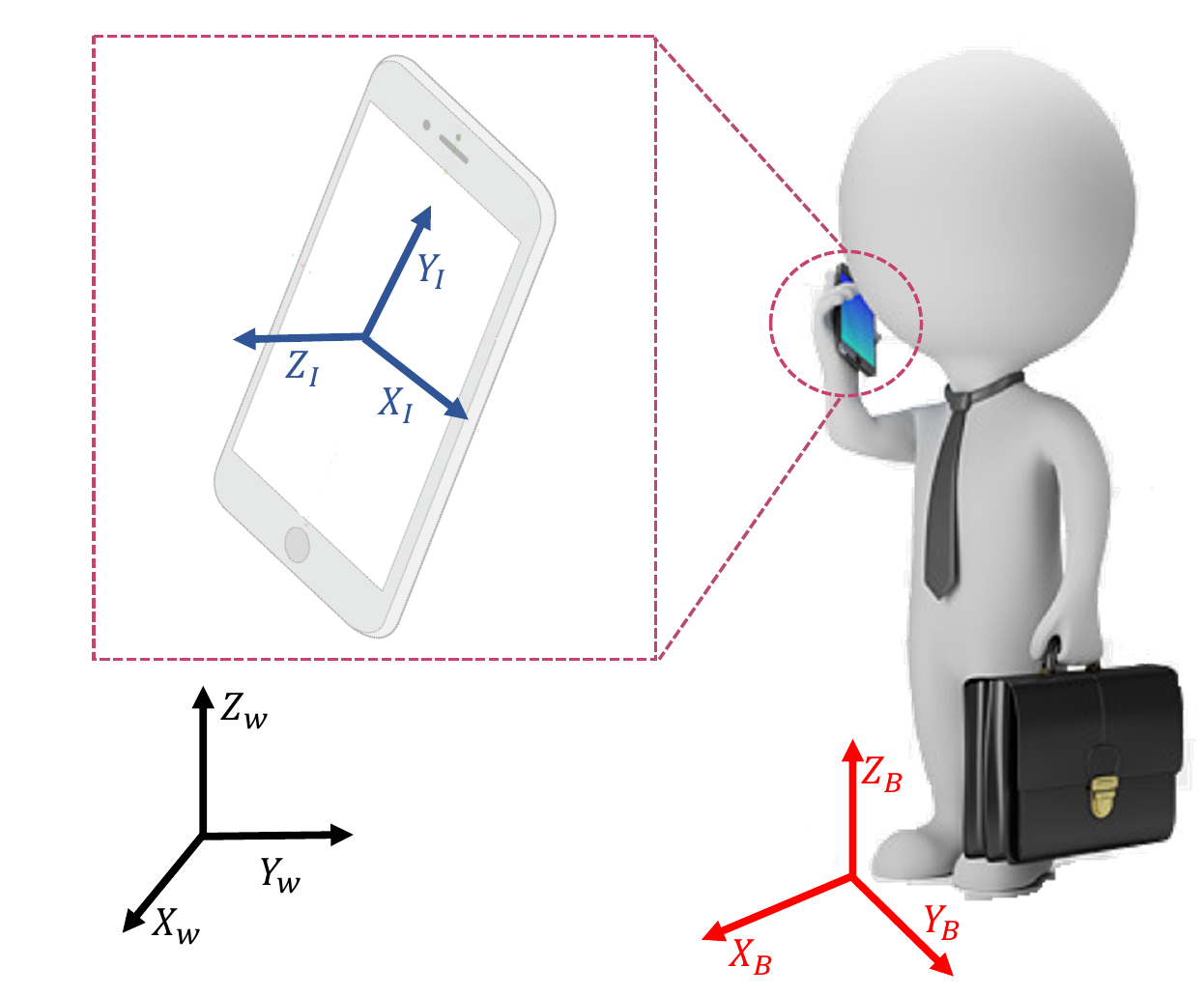}
	\caption{Three coordinate frames used in this paper: W is the world frame (black), I represents the IMU frame (Blue), which is fixed with the phone, and B stands for the ideal body frame (red). }	
	\label{frame}
\end{figure}

IMU double integration for motion estimation has always been attractive to people. First, IMU is energy efficient. It is equipped with every phone and capable of running a whole day without much battery consumption. Second, IMU works anywhere without any prior regional knowledge needed, and the tracking result would not be affected by the environments. The idea for IMU motion estimation is simple, given the orientation by the Android platform, subtracting the gravity from acceleration, integrating the residual acceleration once to get velocity, integrating again to obtain the position. However, small drift in acceleration would result in huge bias in position, in fact, a bias of ${\sigma}$ has an impact of ${\sigma}t^2/2$ in position after t seconds. High precision navigation systems, like a million-dollar military-grade IMU, achieve very small errors, but are too costly in daily use.

Based on the observations that human motions are repetitive and periodic \cite{ormoneit2001learning}, there are various IMU bias correction methods. For example, the widespread Zero velocity UpdaTe (ZUPT) \cite{yun2007self}, in which the IMU is attached on the foot, and then the velocity is corrected to zero each time the foot is connected to the ground. In recent years, due to the rapid development of deep learning and the computation ability of hardware, learning-based methods are gaining much more interest in inertial navigation. In \cite{yan2018ridi}, Yan et al. proposed a robust IMU double integration (RIDI) method, in which the velocity is generated by using support vector regression according to different phone placements. In \cite{yan2019ronin} and \cite{chen2018ionet}, recurrent neural networks are used for end-to-end inertial navigation. 

In this paper, we propose to only utilize the IMU of the phone to realize the pedestrian motion tracking. The Invariant Extended Kalman Filter (IEKF) \cite{barrau2016invariant} and deep learning-based measurement noise adapter are used to reduce the noise in linear acceleration. The main contributions of this paper are summarized as follows:

1. We propose the ideal body coordinate frame (Fig. \ref{frame}) to represent the heading direction, to avoid different estimates of the same motion because of changing the smartphone's holding methods.

2. Our method takes advantage of the IEKF and the learning-based measurement noise adapter. The observation of the IEKF are designed as 3 scalars (velocity in the body frame).

3. Our method is trained by a single object's motion data, and estimates the position and orientation accurately even on the public dataset, in which the data is collected by different object and device.

The remainder of this paper is organized as follows:
We first describe the related methods on IMU model and IEKF model in Sec. \ref{related work}. The system framework is declared in Sec. \ref{system_design}. The implementation details of the system as well as the results are stated in Sec. \ref{implementation_details}, and \ref{results_}, respectively. Finally, the conclusion and future work of this paper are stated in Sec. \ref{conclusions}.

\section{Related Work}  \label{related work}

An IMU is usually a combination of the gyro, the acceleration and magnetometer. Given a fixed coordinate frame and the platform's initial configuration (${\bm{R}^i_0}$, ${\bm{v}^i_0}$, ${\bm{p}^i_0}$), the following orientation ${\bm{R}^i_n}$, velocity ${\bm{v}^i_n}$ and position ${\bm{p}^i_n}$ could be calculated by using the angular velocity and acceleration. Kalman Filter is one of the most widely used methods in motion estimation because it is both extremely simple and general, and we introduce the state-of-the-art IEKF in this section.


\subsection{IMU model}
In this paper, we use the linear acceleration provided by the smartphone, whose gravity has already been removed from the acceleration. The output linear acceleration ${\bm{a}^i_n}$ and the  angular velocity ${\bm{\omega}^i_n}$ are in response to the true linear acceleration ${\overline{\bm{a}}^i_n}$ and the  true angular velocity ${\overline{\bm{\omega}}^i_n}$ \cite{kok2017using}:
\begin{align}
\label{acce}
 & \bm{a}^i_n  = \overline{\bm{a}}^i_n + \bm{b}^a_n + \bm{w}^a_n   \\[1mm]
 \newline
\label{omega}
 & \bm{\omega}^i_n  = \overline{\bm{\omega}}^i_n + \bm{b}^{\omega}_n + \bm{w}^{\omega}_n
\end{align}
The $\bm{b}^a_n$ and $\bm{b}^{\omega}_n$ are constant biases, $\bm{w}^a_n$ and $\bm{w}^{\omega}_n$ are zero-mean Gaussian noises. The bias $\bm{b}^a_n$ and $\bm{b}^{\omega}_n$ follow random walks:
\begin{align}
\label{bacce}
 & \bm{b}^a_{n+1} = \bm{b}^a_n  + \bm{w}^{w^{b_w}}_n   \\[1mm]
\label{bomega}
 & \bm{b}^{\omega}_{n+1}  = \bm{b}^{\omega}_n +  +  \bm{w}^{w^{b_a}}_n
\end{align}
Where $\bm{w}^{w^{b_w}}_n$ and $\bm{w}^{w^{b_a}}_n$ are zero-mean Gaussian noise. Using the Lie group \cite{barrau2016invariant}, the forward kinematic functions are:
\begin{align}
\label{Rpro}
 & \bm{R}_{n+1}^i = \bm{R}_n^i exp((\overline{\bm{\omega}}_idt)_{\times})   \\[1mm]
 \label{ppro}
 & \bm{v}_{n+1}^i = \bm{v}_n^i + \bm{R}_n^i\overline{\bm{a}}_idt   \\[1mm]
 \label{ppropagate}
 & \bm{p}_{n+1}^i = \bm{p}_n^i + \bm{v}_n^idt
\end{align}
The $(\cdot)_{\times}$ in equation (\ref{Rpro}) represents the skew-symmetric matrix associated with the cross product of a 3-dimensional vector. Specifically, in our paper, ${\bm{R}^i_n}$ is the IMU frame with respect to the world frame, ${\bm{v}^i_n}$ and ${\bm{p}^i_n}$ are velocity and position of the IMU in the world frame.

\subsection{Invariant Extended Kalman Filter (IEKF)}

Extended Kalman Filter (EKF) is designed for bias correction from a series of measurements with statistical noise and other inaccuracies, which is calculated by linearizing a non-linear function around the average of the current state. The algorithm contains two-step processes: prediction and update step. In the first step, the next step variables with their uncertainties are estimated according to the current state and the measured input variable $\bm{u}_n$:
\begin{align}
\label{fpropa}
& \hat{\bm{x}}_{n+1} = f(\bm{x}_n, \bm{u}_n) + \bm{\omega}_n   \\[1mm]
\newline
\label{Ppro}
& \hat{\bm{P}}_{n+1} = \bm{F}_n\bm{P}_n{\bm{F}_n}^T+\bm{G}_n\bm{Q}_n{\bm{G}_n}^T  
\end{align}
The $\bm{w}_n$ is the process noise and is assumed to be zero-mean Gaussian noise with covariance matrix $\bm{Q}_n$. The covariance matrix $\bm{P}_n$ is the uncertainty associate with the state $\bm{x}_n$. The $\bm{F}_n$ and $\bm{G}_n$ are Jacobian matrices of $f(\cdot)$ with respect to $\bm{x}_n$ and $\bm{u}_n$.

The second step is to update the next time state ${\bm{x}}_{n+1}$ based on the observation function:
\begin{equation}
\hat{\bm{y}}_{n+1} = h(\hat{\bm{x}}_{n+1}) + \bm{n}_{n+1}
\label{Ofun}
\end{equation}
Where $h(\cdot)$ is the observed function, and the $\bm{n}_{n+1}$ is the observation noise which is assumed to be zero-mean Gaussian white noise with covariance $\bm{R}_n$: $n_n \sim \mathcal{N} (0, \bm{N}_n)$. The covariance matrix $\bm{N}_n$ is set by the user, the smaller $\bm{N}_n$ respects the higher accuracy of the $\hat{\bm{x}}_{n+1}$ in equation (\ref{fpropa}). 

The following steps are computing the Kalman gain $\bm{K}$ and update the state and its covariance matrix:
\begin{align}
\label{k}
& \bm{K} = \hat{\bm{P}}_{n+1}\bm{H}^T{(\bm{H}\hat{\bm{P}}_{n+1}\bm{H}^T+\bm{R})}^{-1}   \\[1mm]
\newline
\label{xcor}
& \bm{x}_{n+1} = \hat{\bm{x}}_{n+1} +\bm{K}(\bm{y}_{n+1}-\hat{\bm{y}}_{n+1})   \\[1mm]
\newline
\label{Pcor}
&\bm{P}_{n+1}=(\bm{I}-\bm{K}\bm{H})\hat{\bm{P}}_{n+1}
\end{align}
Where $\bm{H}$ is the Jacobian of $h(\cdot)$ with respect to $\bm{x}_n$, $\bm{y}_{n+1}$ is the measured observations.

\section{System Design}   \label{system_design}
In this section, we will first give an overview of our system, and then introduce the key issues in the whole system design.

\subsection{System Overview}
\begin{figure}[tbp]
	\centering 
	\includegraphics[width=3.2in]{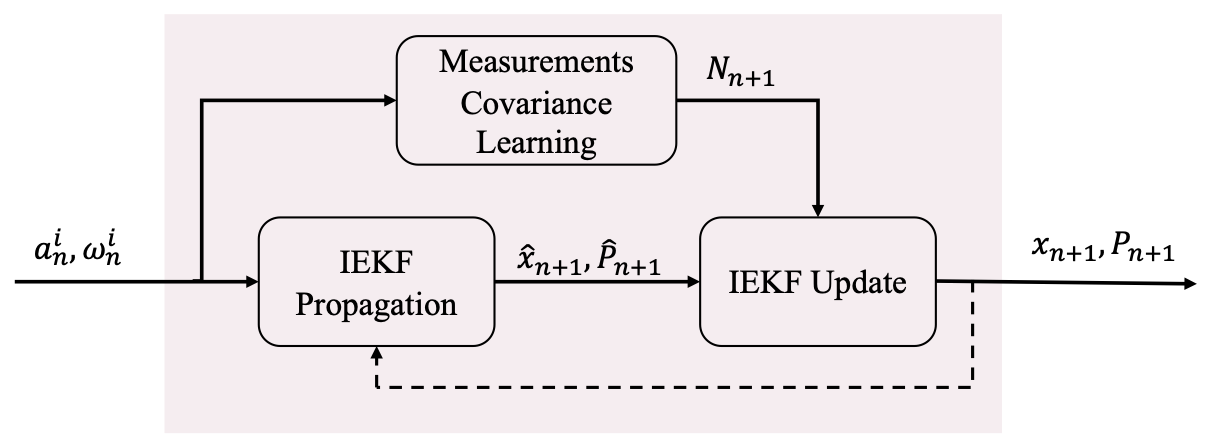}
	\caption{The structure of the whole system. The dotted black line represents the state of $n+1$ is the input of $n+2$. The system yields a real-time estimate of the state $\bm{x}_{n+1}$ along with covariance $\bm{P}_{n+1}$ from raw linear acceleration and angular velocity of the smartphone signal only. }
	\label{overview}
\end{figure}

The main observation of this paper is that human walkings are normally slow and we only focus on planar motions and leave non-planar movements for future work, thus the vertical velocity is null. Based on these assumptions, we use the traditional IMU signal processing model, and IEKF for the bias correction. For the measurement noise matrix in IEKF, a learning-based method is used to adapt with different human walking. The framework of the system is shown as Fig. \ref{overview}.

Given the initial configuration (${\bm{R}^i_0}$, ${\bm{v}^i_0}$, ${\bm{p}^i_0}$), our purpose is to estimate (${\bm{R}^i_n}$, ${\bm{v}^i_n}$, ${\bm{p}^i_n}$) only from the linear acceleration $\bm{a}_n^i$ and angular velocity $\bm{\omega}_n^i$.  We observed that people's daily movements are only in a fixed orientation. As shown in Fig. \ref{frame}, the walking direction is $X_B$ in the body frame, and the velocity along with $Y_B$ and $Z_B$ are zero, thus ${\bm{v}_b = \left[ v_{b_x}, 0, 0 \right]}$ could be used as observations. Our state variables are:
\begin{equation}
\bm{x}_n = (\bm{R}^i_n, \bm{v}^i_n, \bm{p}^i_n, \bm{b}^{\omega}_n, \bm{b}^a_n, \bm{R}^b_{in})
\label{ffun}
\end{equation}
The $\bm{R}^b_{in}$ is the orientation discrepancy of body frame with respect to the IMU frame.

Our state-space is similar as in \cite{brossard2020ai}, except $\bm{R}^b_{in}$. The system in \cite{brossard2020ai} is designed for motion estimation of car. The difference between human tracking and \cite{brossard2020ai} is that in the car tracking system, the IMU is fixed to the car, and the orientation misalignment between the car frame and the IMU frame approximately equals to a three-dimensional identity matrix. The forward direction of the car could also roughly be treated as of the IMU moving direction, this is also why the IMU has null lateral and vertical velocities in the car frame. In our indoor tracking problem, the orientation from IMU frame to the body frame is changeable, and the IMU's velocity expressed in the body frame is also dynamic and non-zero in lateral or upward direction.

\subsection{IEKF propagation}
The propagations of $(\bm{R}^i_n, \bm{v}^i_n, \bm{p}^i_n, \bm{b}^{\omega}_n, \bm{b}^a_n)$ are already given by equations (\ref{bacce}) to (\ref{ppropagate}). The floor of indoor building is usually parallel to the horizontal plane, thus, the $z$ axes of the body frame and the world frame are in the same direction. The initial velocity in the world frame ${\bm{v}^i_0 = \left[ {{v}^i_0}_x, {{v}^i_0}_y, {{v}^i_0}_z \right]}$ is given, and the initial velocity in the body frame ${\bm{v}_b}_0$ could be calculated:
\begin{align}
 \label{rb0}
 {\bm{v}_b}_0 & = {\bm{R}^b_0}^T\bm{v}^i_0   \\[1mm]
 \label{vb0}
                     & = \left[ \sqrt{{{v}^i_0}^2_x+ {{v}^i_0}^2_y}, 0, {{v}^i_0}_z \right]
\end{align}

According to equation (\ref{rb0}) and (\ref{vb0}), the initial rotation from the body frame to the world frame $\bm{R}^b_0$ could be obtained. Then $\bm{R}^b_{i0}$ could also be calculated according to property of continuous rotation \cite{siciliano2010robotics}:
\begin{equation}
\bm{R}^b_{i0} = {\bm{R}^i_0}^T\bm{R}^b_0   
\label{rbi0}
\end{equation}

We observed that the orientation misalignment between the body frame and the smartphone-based IMU frame is an approximate constant in a period of time due to the repeatability and periodic of both human motion and behavior. Thus, we can model the noise of $\bm{R}^b_{in}$ as a zero-mean Gaussian noise:
\begin{equation}
\bm{R}^b_{i(n+1)}=  \bm{R}^b_{in}exp((\bm{\omega}^{\bm{R}^b}_{n})_{\times})
\label{pbipro}
\end{equation}

Now we have decided the $f(\cdot)$ of equation (\ref{fpropa}), and the process covariance matrix $\bm{Q}_n$ is set as a constant matrix. The Jacobians $\bm{F}_n$ and $\bm{G}_n$ could also be calculated. For the rotation matrix propagation function (\ref{Rpro}) and (\ref{pbipro}), we use the first-order Taylor expanx on $exp(\bm{\omega}t)=\bm{I}+\bm{\omega} t$ to realize the derivate. 
\begin{equation}
\bm{F}_n= \bm{I}_{18\times 18}+
\left[ 
\begin{matrix}
\bm{0}     &    \bm{0}    &    \bm{0}     &   -\bm{R}^i_n                                   &    \bm{0}           &    \bm{0}    \\
\bm{0}     &    \bm{0}    &    \bm{0}     &   -(\bm{v^i_n})_{\times}\bm{R}^i_n    &    -\bm{R}^i_n    &    \bm{0}    \\
\bm{0}     &    \bm{I}_{3\times 3}    &    \bm{0}     &   -(\bm{p^i_n})_{\times}\bm{R}^i_n    &    -\bm{0}    &    \bm{0}    \\
\bm{0}     &    \bm{0}    &    \bm{0}     &   -\bm{0}                                  &    \bm{0}           &    \bm{0}    \\                          
\end{matrix}
\right]
\label{Fn}
\end{equation}

\begin{equation}
\bm{G}_n=
\left[ 
\begin{matrix}
\bm{R}^i_n      &    \bm{0}    &    \bm{0}_{3 \times 9}   \\
(\bm{v^i_n})_{\times}\bm{R}^i_n     &   \bm{R}^i_n    &    \bm{0}_{3 \times 9}    \\
(\bm{p^i_n})_{\times}\bm{R}^i_n     &    \bm{0}   &    \bm{0}_{3 \times 9}    \\
\bm{0}     &    \bm{0}    &    \bm{0}_{3 \times 9}    \\                          
\end{matrix}
\right]
\label{Gn}
\end{equation}


\subsection{Pseudo-Measurements}

The velocity in the body frame could be expressed as: 
\begin{equation}
\bm{v}_{bn}=  {\bm{R}^i_n\bm{R}^b_{in}}^T\bm{v}^i_n
\label{vbn}
\end{equation}

We could further determine the Jacobian of observed function:
\begin{equation}
\bm{H}= \left[ 
\begin{matrix}
\bm{0}  &   {\bm{R}^i_n\bm{R}^b_{in}}^T  &  \bm{0}_{3 \times 12}
\end{matrix}
\right]
\label{Hfun}
\end{equation}

\begin{figure}[tbp]
	\centering 
	\includegraphics[width=3.2in]{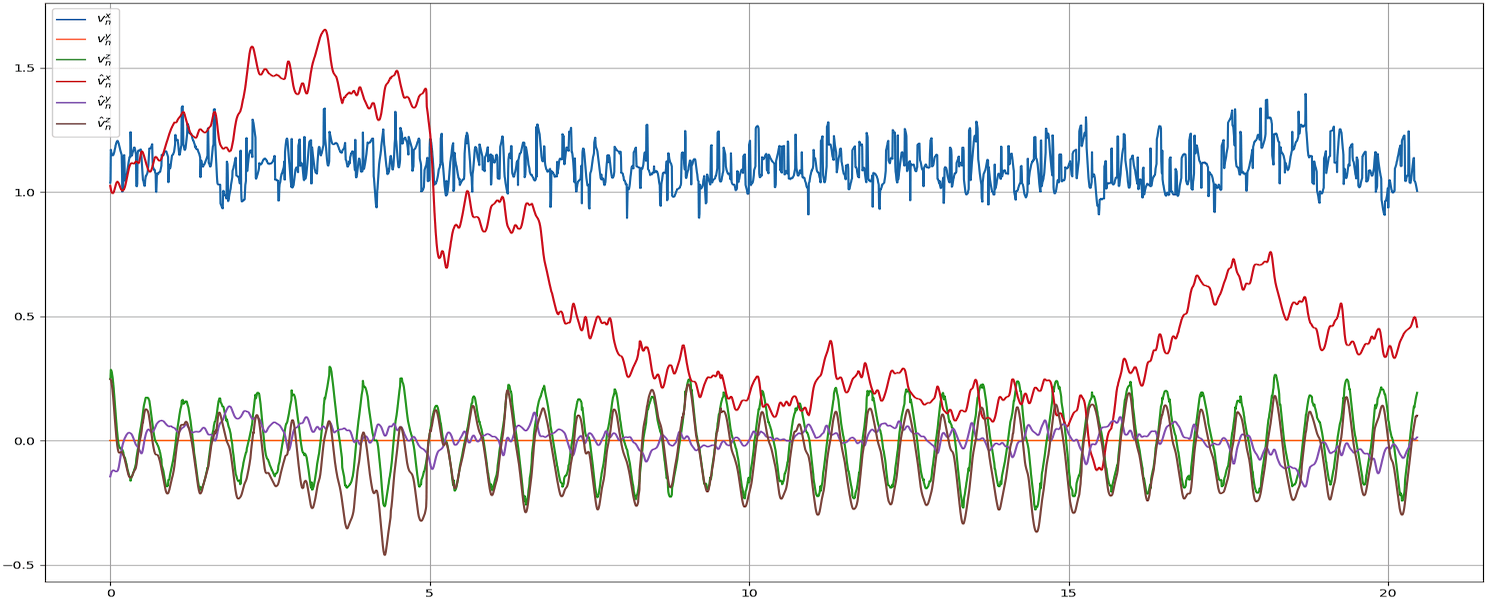}
	\caption{The velocity in the body frame without observation in $\bm{v}_{bx}$. The ground truth of the velocity along with $x$ axis (blue line) is a quasi-constant, while the calculated velocity (red line) deviated a lot in only $20s$. }	
	\label{pseudon}
\end{figure}

In \cite{brossard2020ai}, the null lateral and vertical velocities are used as observations, while there is no constraint in the forward direction. The method works fine in the car inertial navigation system, but in our smartphone-based tracking, the lateral and vertical constraints make limited effect to the forward direction. As shown in Fig. \ref{pseudon}, the forward velocity would drift a lot within the first 5 seconds.  

In this paper, except for the null lateral and vertical velocity, we utilize the periodic and slowness of human walking, which indicates that human motions are relatively slow and smooth. We treat the average speed of the past 5 seconds as the current speed:
\begin{equation}
\bm{v}_{b(n+1)x}=   \sqrt{(\bm{p}^i_n-\bm{p}^i_{n-win\ast 5})/5}
\label{vbno}
\end{equation}
Where $win$ in (\ref{vbno}) is the sampling rate. Moreover, we utilize the constraints of human movements. In daily life, the walking velocity is usually slower than $2 m/s$. Thus, we have  generated 3-scalar pseudo-observations:
\begin{equation}
 \bm{y}_{n+1} = \left[
 \begin{matrix}
 \bm{v}_{b(n+1)x}  &  \bm{0}   &  \bm{0}
 \end{matrix}
 \right]
\end{equation}      

\subsection{Learning-based Measurement Noise Parameter Adapter}
 
As shown in the equation (\ref{Ofun}), the covariance $\bm{N}_n$ is used at each instant $n$ for the filter update. The measurement noise parameter adapter dynamically computes covariances meant to improve the localization accuracy by just using the linear acceleration $\bm{a}^i_n$ and angular velocity $\bm{\omega}^i_n$. The main component of the adapter is a Convolutional Neural Network (CNN) \cite{schmidhuber2015deep}, whose input is a window of $M$ $\bm{a}^i_n$ and $\bm{\omega}^i_n$. A relatively simple network with small number of parameters is used to avoid over-fitting and made its output independent of the state estimates.
 
The total components of the adapter consist of several full layers followed by a full layer outputting a vector $\bm{Z}_n={3\ast\left[tanh(z^{fw)}_n, tanh(z^{lat}_n), tanh(z^{up}_n) \right]}$. The output covariance matrix $\bm{N}_{n+1}$ is then computed by:
 \begin{equation}
 \bm{N}_{n+1}=  diag(\sigma^2_{fw}10^{\bm{Z}_n\left[1\right]}, \sigma^2_{lat}10^{\bm{Z}_n\left[2\right]}, \sigma^2_{up}10^{\bm{Z}_n\left[3\right]} )
 \label{Nn1}
 \end{equation}
 Where $\sigma^2_{fw}$, $\sigma^2_{lat}$ and $\sigma^2_{up}$ are our initial guesses for the measurement noise parameters. The output of the network may expand the covariance to $1000\sigma^2$ and compress it up to $10^{-3}\sigma^2$.
 
\section{Implementation Details}     \label{implementation_details}

We provide the settings and implementation details of our method in this section. The inertial signals are collected by smartphone. The latter process is implemented in Python with Pytorch \cite{paszke2019pytorch} library.

\subsection{Data collection and preprocessing}

\begin{figure}[tbp]
	\centering 
	\includegraphics[width=3.3in]{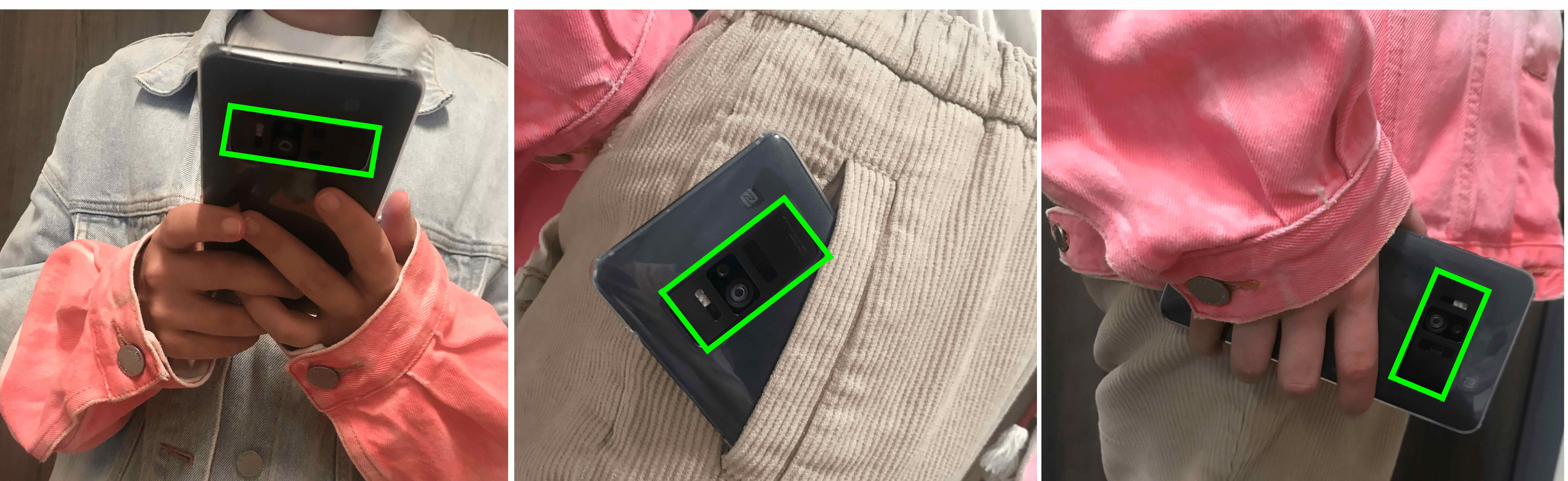}
	\caption{Different phone holding methods in daily life,  while exposing the camera (green box) all the time to collect the ground-truth motions by visual-inertial odometry. The configurations from left to right are: playing while walking; putting in the pocket and holding in the hand.}	
	\label{phoneholding}
\end{figure}

We use a Google Tango phone, ASUS Zenfone AR, to record the $\bm{a}^i_n$, $\bm{\omega}^i_n$, and 3D camera poses. We make sure that the camera has not been blocked all the time (see Fig. \ref{phoneholding}). The signals we used for training are from the same person with different motions.

The sampling rate of the collected data is 200Hz. Asynchronous signals from various sensors are synchronized into the timestamps of Tango poses by using linear interpolation, which is a preprocessing step of our system. The method of collecting the data is related to RIDI dataset, but the constraints of our phone holding methods are looser, all the three phone carrying manners are daily, instead of strapped tightly to the body. We also use the RIDI dataset to verify our system.

\subsection{System Settings}

The initial parameters of IEKF are as follows:
\begin{align}
\label{p1}
\bm{P}_0 & =  diag{(\sigma^{\bm{R}}_0\bm{I}, \sigma^{\bm{v}}_0\bm{I}, \bm{0}_3, \sigma^{\bm{b}^\omega}_0\bm{I}, \sigma^{\bm{b}^a}_0\bm{I}, \sigma^{\bm{R}^b}_0\bm{I})}^2   \\[1.5mm]
 & = diag(\tiny{10^{-6}}\bm{I}, 10^{\tiny{-5}}\bm{I}, \bm{0}_3, 10^{\tiny{-6}}\bm{I}, 10^{\tiny{-3}}\bm{I}, 10^{\tiny{-5}}\bm{I})  \\[1.5mm]
 \label{q1}
 \bm{Q}_0 & = diag{(\sigma_\omega\bm{I}, \sigma_a\bm{I}, \sigma_{\bm{b}^\omega}\bm{I}, \sigma_{\bm{b}^a}\bm{I}, \sigma_{\bm{R}^b}\bm{I})}^2  \\[1.5mm]
 & = diag(2\cdot10^{-4}\bm{I}, 10^{-3}\bm{I}, 10^{-6}\bm{I}, 2\cdot10^{-5}\bm{I}, 10^{-3}\bm{I})   \\[1.5mm]
 \label{n0}
 \bm{N}_0 & = diag(\sigma^2_{fw}, \sigma^2_{lat}, \sigma^2_{up} ) = diag(3, 2, 0.2)
\end{align}  
Where $\bm{I}$ stands for $\bm{I}_{3\times3}$. Except for the 3 scalars output layer, the adapter was designed as a 1D temporal convolutional neural network with 2 layers. To obtain a measurement covariance, we want to it is related to last 20 frames of the $\bm{a}^i_n$, $\bm{\omega}^i_n$. Thus, for the first layer, $in\_channels = 6$, $out\_channels = 32$, $kernel\_size = 6$, $dilation = 2$, while for the second layer, $in\_channels = 32$, $out\_channels = 32$, $kernel\_size = 5$, $dilation = 3$.  

\subsection{Training}

We use a batch size of 400, and for each single batch, six 40-second sequences (8000 sampling timestamps) are sampled. The minimum length of the six sequences is 28000. During training, we use an Adam optimizer with an initial learning rate of 0.001 and the minimum learning rate of $10^{-5}$. For any CNN layer, we apply dropout with the keep probability of 0.5. The loss is determined by:
\begin{equation}
loss =  \frac{1}{8000}\sum_{i}^{8000}{\|\bm{v}_{bi}-\bm{v}_{b\_gti}\|}^2
\label{Nn}
\end{equation}
Where $\bm{v}_{b\_gti}$ is the ground-truth velocity in the body frame. Given the 3D camera pose and the timestamp, the velocity in the world frame could be obtained. The velocity and the orientation of the ideal body frame could then be calculated by using equation (\ref{rb0}) and (\ref{vb0}). The $\bm{v}_{bi}$ is the velocity in the body frame calculated from the output of IEKF. 

For the test set, we use three different types of data: the whole sequences used in the training data; the signals from the same object but not in the training set; the data from RIDI dataset, which is from different carrier and device. 

\begin{figure}[tbp]
	\centering 
	\includegraphics[width=3.2in]{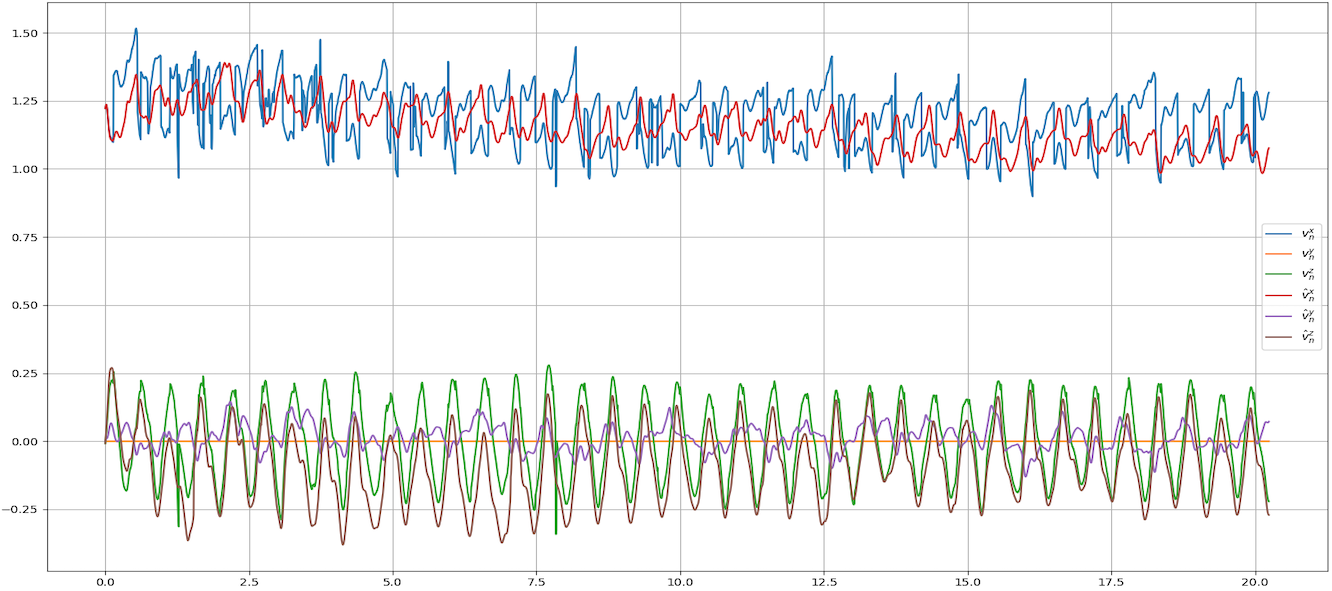}
	\caption{The velocity in the body frame with observation in $\bm{v}_{bx}$. The estimation of the velocity along with $x$ axis (red line) is almost in the same value with the ground truth (blue line), and the discrepancy would be further minimized by the measurement noise adapter.}	
	\label{pseudo}
\end{figure}

\begin{figure*}[tbp]
	\centering
	\subfloat[Seen motion]{\includegraphics[height=2.0in]{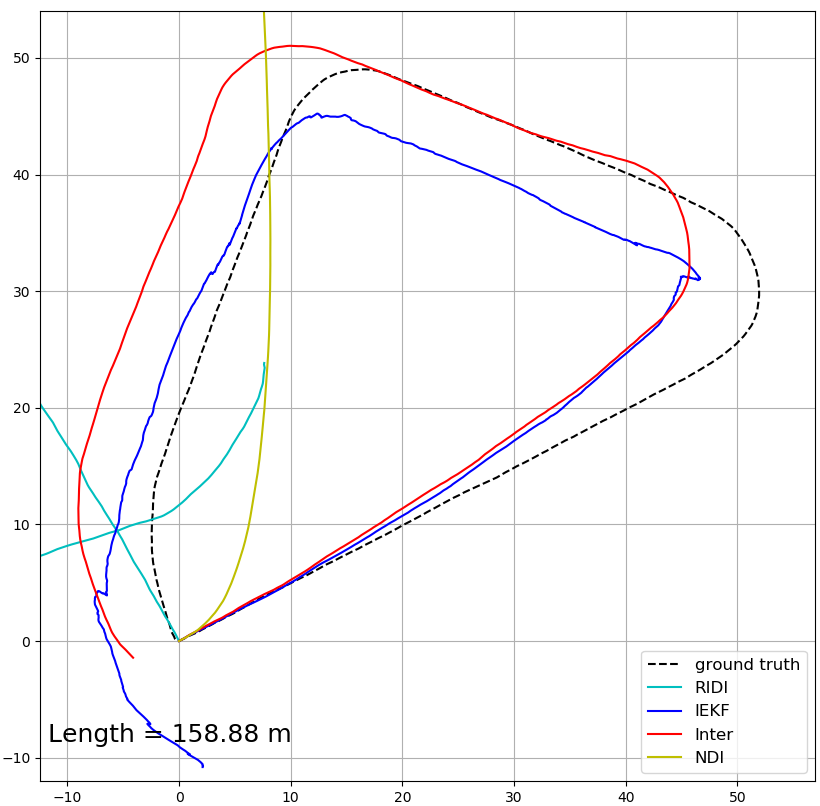}\label{seen}}   \quad
	\subfloat[Unseen motion]{\includegraphics[height=2.0in]{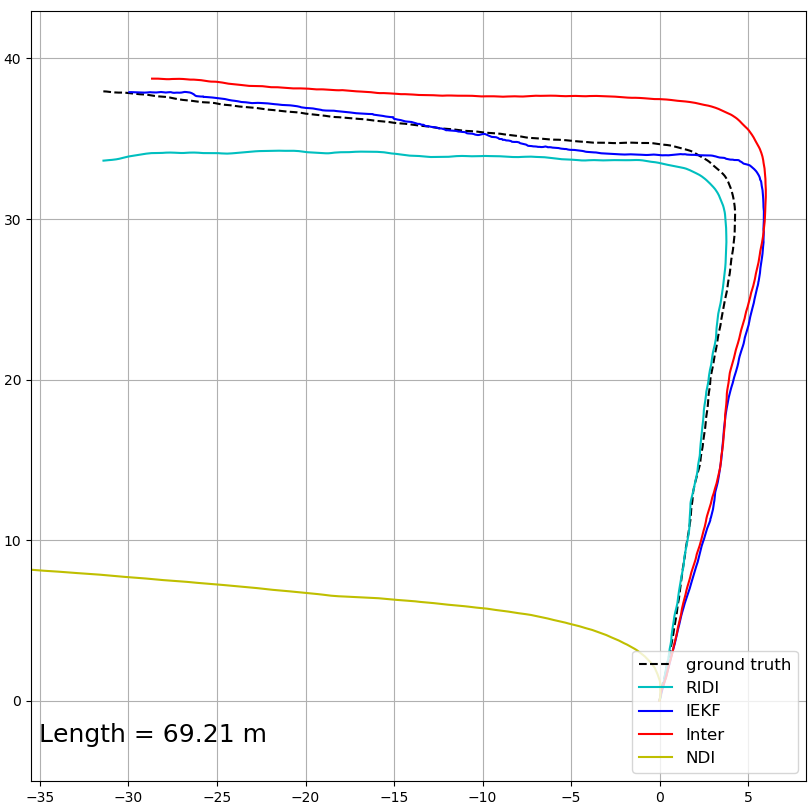}\label{unseen}}   \quad
	\subfloat[RIDI data]{\includegraphics[height=2.0in]{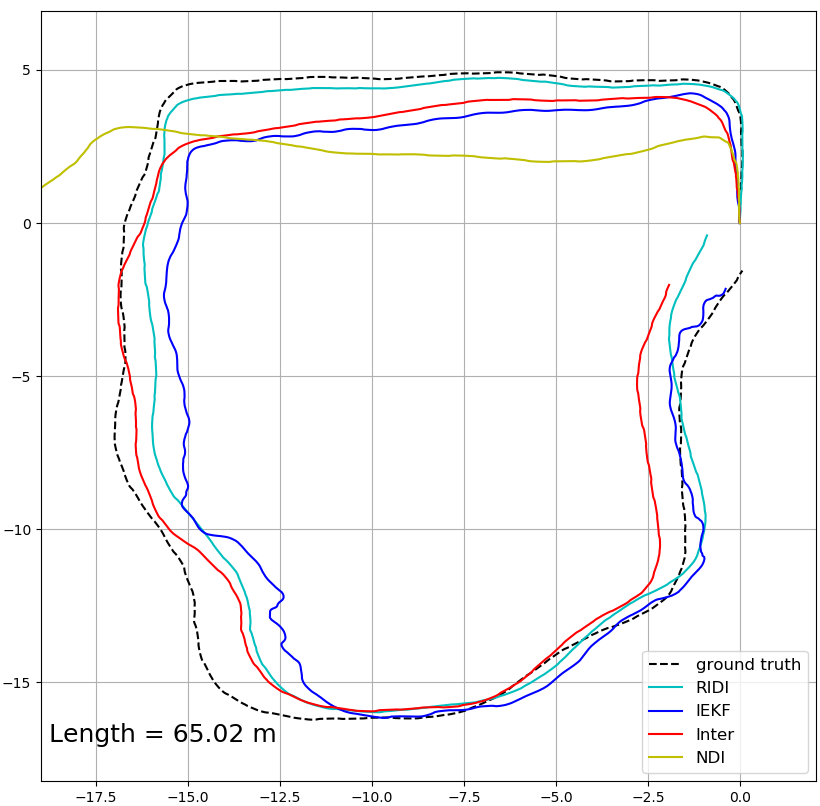}\label{ridi}}
	\caption{Estimated positions: ground truth (black dashed), RIDI (cyan), IEKF (blue), Inter (red) and NDI (yellow).The seen motion is a 145s sequence used in the training data and the phone was playing in the hand while walking. Unseen motion is with the same object as the seen one, but is not contained in the training set, and it is a 65s sequence with the phone holding in the hand. The RIDI data is from the public dataset, which is collected by fixed the phone to the body within 59s.}
	\label{generated p}
\end{figure*}

\section{Results and Analysis}      \label{results_}

In this section, the results of our system are provided, including the pseudo-observations of IEKF, and the position trajectories of three different test datasets (Fig. \ref{generated p}). We compare four methods which are all IMU-based estimations: NDI denotes the naive double integration of the raw linear accelerations; RIDI is the regressed trajectory proposed in \cite{yan2018ridi}, and the training data is also from \cite{yan2018ridi}; IEKF represents the $\bm{p}^i_n$ and Inter is the integration of the $\bm{v}^i_n$ in the state variables.

We use two standard metrics proposed in \cite{sturm2012benchmark}: Absolute Trajectory Error (ATE) defined as the average Root Mean Squared Error (RMSE) between estimated and ground-truth trajectories as a whole; Relative Trajectory Error (RTE) defined as the average RMSE over a fixed time interval, i.e. 1 minute in our evaluation.

\subsection{Pseudo-Observation}

By adding observation in the forward direction, the velocity in the body frame along the $X_B$ axis could be highly corrected. Fig. \ref{pseudo} shows the estimated velocity in the body frame at the started batch. Compared with Fig. \ref{pseudon}, the walking velocity is corrected and the bias is minimized with dynamic measurement covariance based on deep learning. 

\begin{figure}[tbp]
	\centering 
	\includegraphics[width=3.2in]{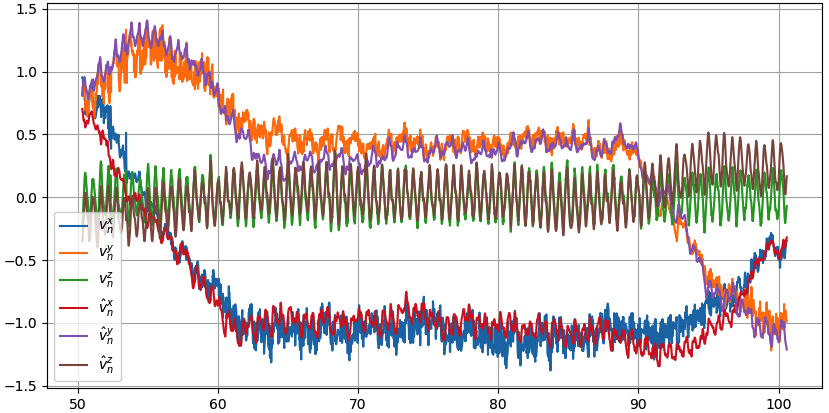}
	\caption{The ground-truth $(v^x_n, v^y_n, v^z_n)$ and recovered $(\hat{v}^x_n, \hat{v}^y_n, \hat{v}^z_n)$ velocity in the world frame of the sequence whose trajectory is shown in Fig. \ref{seen}. The time index is from 50s to 100s, where two turns are contained.}
	\label{velocity}
\end{figure}

\subsection{Seen Motion}

We notice that the estimated velocity $\bm{v}^i_n$, one of the state variables of the IEKF, could be almost accurately calculated, even with two turns within 50 seconds  (see Fig. \ref{velocity}). The estimated $\bm{p}^i_n$ is not as robust as the velocity, and the error drifts over time. Surprisingly, the integration of $\bm{v}^i_n$ performs a better position estimation (IEKF and Inter in Fig. \ref{seen}), in terms of the start and end point, and the calculated RTE (TABLE \ref{metric}). 
 
\subsection{Unseen Motion }

For the sequence not contained in the training set, the estimation of our system is guaranteed, too (Fig. \ref{unseen}). In this simple straight walk with a left turn, the output position estimate of IEKF outperforms Inter, which may come from the fact that Kalman Filter is more likely to converge within small noise.

\begin{table}[htbp]
	
	\caption{Positional accuracy evaluations (in meters)}
	
	\label{metric}
	
	\begin{center}
		
		\begin{tabular}{|c|c|c|c|c|c|}
			
			\hline
			
			Metric & Method  &Seen  &  Unseen  &RIDI data  &Averaged \\
			
			\hline
			\multirow{4}{*}{ATE}     &NDI    &446.80  &149.52  &17.44  &204.59\\   \cline{2-6}		 
			& RIDI &41.95  &1.24 &0.77  &14.65  \\   \cline{2-6}		
			& IEKF &3.92  &0.91  &1.28  &2.04  \\   \cline{2-6}		
			& Inter &3.96  &1.68  &0.95  &2.20  \\
			\hline	
			\multirow{4}{*}{RTE}     &NDI    &34.25  &249.11  &33.19  &105.52 \\   \cline{2-6}		      
			& RIDI &52.27  &1.80  &0.90  &18.32  \\     \cline{2-6}		
			& IEKF &3.63  &0.85  &0.28  & 1.59  \\    \cline{2-6}
			& Inter &2.85  &1.76  &0.98  &1.87  \\
			\hline	
		\end{tabular}
		
	\end{center}
	
\end{table}

\subsection{Data in RIDI Dataset}

The test results of RIDI data is shown as Fig. \ref{ridi}, both the IEKF and Inter methods perform well towards the RIDI data, especially the RTE of IEKF (0.28\textit{m}) is much lower than RIDI method itself (0.90\textit{m}), whose training data is from the same device as the test set. Besides, when integrating the raw data twice directly, the ATE of RIDI data is much smaller than our data, which also indicates the extensiveness of our data and the robustness of our method. 

From the tests, and the average error of position from IEKF and Inter, we conclude that when the sequence is within 1 min, the IEKF could provide accurate position estimation, but when the time is longer, the Inter method would be more accurate, which may benefit from the velocity in the body frame as the loss function.

\section{Conclusion and Future Work}     \label{conclusions}

 We propose a novel approach for inertial pedestrian dead reckoning in this paper. The neural network is exploited to dynamically adapt the covariance of the EKF that performs localization, velocity and sensor bias estimation. The whole system is fed with the IMU signals collected by the smartphone only. The method leads to surprisingly accurate results not only for the self-collected data but for the data in public dataset. The position state variable has a higher accuracy within 1-minute walking, while the integration of the velocity variable outperforms in longer walkings. In the future, we may take advantage of these two variables to further improve positional accuracy.
 


\section{Acknowledgement}

This project is partially supported by Hong Kong ITC ITSP Tier 2 grant \#ITS/105/18FP: An intelligent Robotics System for Autonomous Airport Passenger Trolley Deployment awarded to Max Q.-H. Meng.

\normalem
\bibliographystyle{IEEEtran}
\bibliography{iros2020.bib}

\begin{thebibliography}{10}
\providecommand{\url}[1]{#1}
\csname url@rmstyle\endcsname
\providecommand{\newblock}{\relax}
\providecommand{\bibinfo}[2]{#2}
\providecommand\BIBentrySTDinterwordspacing{\spaceskip=0pt\relax}
\providecommand\BIBentryALTinterwordstretchfactor{4}
\providecommand\BIBentryALTinterwordspacing{\spaceskip=\fontdimen2\font plus
\BIBentryALTinterwordstretchfactor\fontdimen3\font minus
  \fontdimen4\font\relax}
\providecommand\BIBforeignlanguage[2]{{%
\expandafter\ifx\csname l@#1\endcsname\relax
\typeout{** WARNING: IEEEtran.bst: No hyphenation pattern has been}%
\typeout{** loaded for the language `#1'. Using the pattern for}%
\typeout{** the default language instead.}%
\else
\language=\csname l@#1\endcsname
\fi
#2}}

\bibitem{langlois2017indoor}
C.~Langlois, S.~Tiku, and S.~Pasricha, ``Indoor localization with smartphones:
  Harnessing the sensor suite in your pocket,'' \emph{IEEE Consumer Electronics
  Magazine}, vol.~6, no.~4, pp. 70--80, 2017.

\bibitem{cheung2005multidimensional}
K.~W. Cheung and H.-C. So, ``A multidimensional scaling framework for mobile
  location using time-of-arrival measurements,'' \emph{IEEE transactions on
  signal processing}, vol.~53, no.~2, pp. 460--470, 2005.

\bibitem{shen2010fundamental}
Y.~Shen and M.~Z. Win, ``Fundamental limits of wideband localization—part i:
  A general framework,'' \emph{IEEE Transactions on Information Theory},
  vol.~56, no.~10, pp. 4956--4980, 2010.

\bibitem{zimmermann2012gsm}
L.~Zimmermann, A.~Goetz, G.~Fischer, and R.~Weigel, ``Gsm mobile phone
  localization using time difference of arrival and angle of arrival
  estimation,'' in \emph{International Multi-Conference on Systems, Signals \&
  Devices}.\hskip 1em plus 0.5em minus 0.4em\relax IEEE, 2012, pp. 1--7.

\bibitem{yang2015wifi}
C.~Yang and H.-R. Shao, ``Wifi-based indoor positioning,'' \emph{IEEE
  Communications Magazine}, vol.~53, no.~3, pp. 150--157, 2015.

\bibitem{wang2016indoor}
B.~Wang, Q.~Chen, L.~T. Yang, and H.-C. Chao, ``Indoor smartphone localization
  via fingerprint crowdsourcing: Challenges and approaches,'' \emph{IEEE
  Wireless Communications}, vol.~23, no.~3, pp. 82--89, 2016.

\bibitem{liang2018indoor}
Q.~Liang, L.~Wang, Y.~Li, and M.~Liu, ``Indoor mapping and localization for
  pedestrians using opportunistic sensing with smartphones,'' in \emph{2018
  IEEE/RSJ International Conference on Intelligent Robots and Systems
  (IROS)}.\hskip 1em plus 0.5em minus 0.4em\relax IEEE, 2018, pp. 1649--1656.

\bibitem{hesch2014camera}
J.~A. Hesch, D.~G. Kottas, S.~L. Bowman, and S.~I. Roumeliotis,
  ``Camera-imu-based localization: Observability analysis and consistency
  improvement,'' \emph{The International Journal of Robotics Research},
  vol.~33, no.~1, pp. 182--201, 2014.

\bibitem{qin2018vins}
T.~Qin, P.~Li, and S.~Shen, ``Vins-mono: A robust and versatile monocular
  visual-inertial state estimator,'' \emph{IEEE Transactions on Robotics},
  vol.~34, no.~4, pp. 1004--1020, 2018.

\bibitem{WinNT}
``{Google. project Tango},''
  \url{https://en.wikipedia.org/wiki/Tango_(platform)}.

\bibitem{ormoneit2001learning}
D.~Ormoneit, H.~Sidenbladh, M.~J. Black, and T.~Hastie, ``Learning and tracking
  cyclic human motion,'' in \emph{Advances in Neural Information Processing
  Systems}, 2001, pp. 894--900.

\bibitem{yun2007self}
X.~Yun, E.~R. Bachmann, H.~Moore, and J.~Calusdian, ``Self-contained position
  tracking of human movement using small inertial/magnetic sensor modules,'' in
  \emph{Proceedings 2007 IEEE International Conference on Robotics and
  Automation}.\hskip 1em plus 0.5em minus 0.4em\relax IEEE, 2007, pp.
  2526--2533.

\bibitem{yan2018ridi}
H.~Yan, Q.~Shan, and Y.~Furukawa, ``Ridi: Robust imu double integration,'' in
  \emph{Proceedings of the European Conference on Computer Vision (ECCV)},
  2018, pp. 621--636.

\bibitem{yan2019ronin}
H.~Yan, S.~Herath, and Y.~Furukawa, ``Ronin: Robust neural inertial navigation
  in the wild: Benchmark, evaluations, and new methods,'' \emph{arXiv preprint
  arXiv:1905.12853}, 2019.

\bibitem{chen2018ionet}
C.~Chen, X.~Lu, A.~Markham, and N.~Trigoni, ``Ionet: Learning to cure the curse
  of drift in inertial odometry,'' in \emph{Thirty-Second AAAI Conference on
  Artificial Intelligence}, 2018.

\bibitem{barrau2016invariant}
A.~Barrau and S.~Bonnabel, ``The invariant extended kalman filter as a stable
  observer,'' \emph{IEEE Transactions on Automatic Control}, vol.~62, no.~4,
  pp. 1797--1812, 2016.

\bibitem{kok2017using}
M.~Kok, J.~D. Hol, and T.~B. Sch{\"o}n, ``Using inertial sensors for position
  and orientation estimation,'' \emph{arXiv preprint arXiv:1704.06053}, 2017.

\bibitem{brossard2020ai}
M.~Brossard, A.~Barrau, and S.~Bonnabel, ``Ai-imu dead-reckoning,'' \emph{IEEE
  Transactions on Intelligent Vehicles}, 2020.

\bibitem{siciliano2010robotics}
B.~Siciliano, L.~Sciavicco, L.~Villani, and G.~Oriolo, \emph{Robotics:
  modelling, planning and control}.\hskip 1em plus 0.5em minus 0.4em\relax
  Springer Science \& Business Media, 2010.

\bibitem{schmidhuber2015deep}
J.~Schmidhuber, ``Deep learning in neural networks: An overview,'' \emph{Neural
  networks}, vol.~61, pp. 85--117, 2015.

\bibitem{paszke2019pytorch}
A.~Paszke, S.~Gross, F.~Massa, A.~Lerer, J.~Bradbury, G.~Chanan, T.~Killeen,
  Z.~Lin, N.~Gimelshein, L.~Antiga, \emph{et~al.}, ``Pytorch: An imperative
  style, high-performance deep learning library,'' in \emph{Advances in Neural
  Information Processing Systems}, 2019, pp. 8024--8035.

\bibitem{sturm2012benchmark}
J.~Sturm, N.~Engelhard, F.~Endres, W.~Burgard, and D.~Cremers, ``A benchmark
  for the evaluation of rgb-d slam systems,'' in \emph{2012 IEEE/RSJ
  International Conference on Intelligent Robots and Systems}.\hskip 1em plus
  0.5em minus 0.4em\relax IEEE, 2012, pp. 573--580.

\end{thebibliography}

\end{document}